\documentclass{article}
\usepackage{spconf,amsmath,graphicx,hyperref}
\usepackage{balance}
\usepackage{orcidlink}
\usepackage{float}
\usepackage{amsmath,amsfonts}
\usepackage{algorithmic}
\usepackage{array}
\usepackage[caption=false,font=normalsize,labelfont=sf,textfont=sf]{subfig}
\usepackage{textcomp}
\usepackage{stfloats}
\usepackage{url}
\usepackage{verbatim}
\usepackage{graphicx}

\usepackage[utf8]{inputenc} %
\usepackage[T1]{fontenc}    %
\usepackage{hyperref}       %
\usepackage{url}            %
\usepackage{booktabs}       %
\usepackage{amsfonts}       %
\usepackage[ruled,vlined]{algorithm2e} %
\usepackage{graphicx}       %
\usepackage{nicefrac}       %
\usepackage{microtype}      %
\usepackage{xcolor}         %
\usepackage{multirow}
\usepackage{amssymb}
\usepackage{amsmath}

\usepackage{wrapfig} %
\usepackage{booktabs}  %
\usepackage{lipsum}    %

\title{Robust Multimodal Representation Learning in Healthcare}
\name{Xiaoguang Zhu$^{1*\dagger}$, Linxiao Gong$^{2*}$, Lianlong Sun$^3$, Yang Liu$^{4\dagger}$, Haoyu Wang$^5$, Jing Liu$^{6,7}$ \thanks{{*} Equal contribution. $^\dagger$Corresponding author.}}
  \address{$^1$University of California, Davis \quad $^2$HKUST (GZ) \quad $^3$University of Rochester\quad $^4$ Tongji University  \\ \quad $^5$Georgia Institute of Technology \quad $^6$Fudan University \quad $^7$The University of British Columbia  }

\begin{document}

\maketitle

\begin{abstract}
Medical multimodal representation learning aims to integrate heterogeneous data into unified patient representations to support clinical outcome prediction. However, real-world medical datasets commonly contain systematic biases from multiple sources, which poses significant challenges for medical multimodal representation learning. Existing approaches typically focus on effective multimodal fusion, neglecting inherent biased features that affect the generalization ability. To address these challenges, we propose a Dual-Stream Feature Decorrelation Framework that identifies and handles the biases through structural causal analysis introduced by latent confounders. Our method employs a causal-biased decorrelation framework with dual-stream neural networks to disentangle causal features from spurious correlations, utilizing generalized cross-entropy loss and mutual information minimization for effective decorrelation. The framework is model-agnostic and can be integrated into existing medical multimodal learning methods. Comprehensive experiments on MIMIC-IV, eICU, and ADNI datasets demonstrate consistent performance improvements.

\end{abstract}


\section{Introduction}
\label{sec:intro}

Medical data naturally exhibits multimodal and heterogeneous characteristics \cite{sylvain2021cmim}. To provide accurate diagnoses, healthcare professionals must thoroughly analyze various patient data modalities to support diagnosis and personalized treatment \cite{johnson2023mimic,wagner2020ptb,maltesen2020longitudinal,allen2014uk}. Consequently, multimodal representation learning in healthcare has emerged as a promising paradigm, aiming to integrate information from diverse modalities and predict downstream clinical outcomes \cite{liu2025multimodal}.

To handle this multimodal heterogeneity, researchers have proposed various multimodal representation learning approaches that aim to integrate heterogeneous clinical data into unified patient representations \cite{su2025large}. Existing methods for multimodal representation learning can be broadly grouped into Transformer-based fusion strategies and graph-based approaches. Transformer-based methods~\cite{ma2022multimodal}, encode each modality separately and use cross-attention to model cross-modal dependencies and fuse information into unified representations. Graph-based methods~\cite{chen2020hgmf,zhang2022m3care,wu2024multimodal,LIU2022109697, 10219171}, construct patient-modality graphs and leverage message passing to capture correlations and complementary information across modalities. 

However, current medical multimodal learning methods face a significant challenge, where medical datasets often contain multiple types of systematic biases, such as biases from data acquisition processes and introduced by latent confounding factors. These biases significantly affect the performance and generalizability of medical multimodal models across diverse patient populations. For example, cardiovascular disease prediction models trained predominantly on male datasets lead to significantly increased misdiagnosis rates of heart attacks in women~\cite{cross2024bias}. Similarly, convolutional neural networks for skin lesion classification are primarily trained using images with Black patient data comprising only 5\%-10\%, resulting in significantly lower performance when tested in Black patients \cite{norori2021addressing}. Such cases highlight the urgent need for bias-aware multimodal learning frameworks that can deliver robust and equitable outcomes. Although causal reasoning methods have recently been proposed to mitigate bias in medical learning \cite{deshpande2022deep,yang2021causalvae}, they are primarily designed for single-modality settings and have rarely been extended to multimodal scenarios, particularly those involving modality missingness \cite{liu2025privacy}.

To address these challenges, we propose a \textbf{D}ual-Stream \textbf{F}eature \textbf{D}ecorrelation (\textbf{DFD}) Framework that introduces causal reasoning into data bias handling and systematically separates causal and distributional bias in medical multimodal learning. Specifically, we design a dual-stream causal-biased feature decorrelation module that utilizes adaptive gating mechanisms to decompose graph neural networks into causal and bias streams. The causal stream learns stable features through standard cross-entropy loss, while the bias stream extracts the biased features via generalized cross-entropy loss. Finally, mutual information minimization ensures statistical independence between the two streams, thereby achieving effective decorrelation of causal and biased features. Comprehensive evaluations on MIMIC-IV, eICU, and ADNI datasets demonstrate consistent improvements, validating the effectiveness and practicality of our frameworks. Additionally, our framework is model-agnostic and can be flexibly integrated into existing methods with considerable modifications.

\section{Methodology}

\begin{figure*}[t]
    \centering 
    \includegraphics[width=1\textwidth]{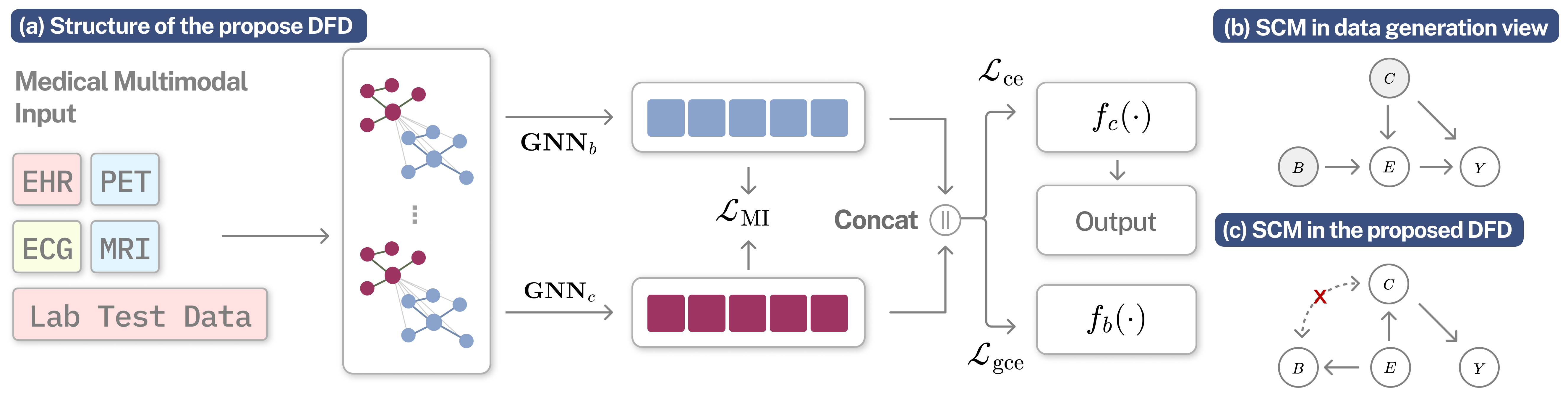}
    \vspace{-10pt}
    \caption{Overview of the proposed DFD framework. (a) The structure of the proposed dual-stream feature decorrelation network. (b) SCM in data generation view. Gray color indicates that the features are not observable. (c) SCM in the proposed method.}
    \label{fig: Overview}
     \vspace{-5pt}
\end{figure*}

\subsection{Causal Analysis}
\label{sec:causal}
We formalize the data generation process and the behavior of the model using the Structured Causal Model (SCM) \cite{pearl2016causal,pearl2009causality,liu2025crcl}. For the conventional multimodal learning methods, they use the deep model to extract the embedding $\mathbf{E}$ from multimodal input. However, the multimodal input may be biased, as the data not only contains the causal features $\mathbf{C}$ that are truly predictive of the outcome $\mathbf{Y}$, but also biased features $\mathbf{B}$ that are spuriously correlated with the outcome due to various factors such as data acquisition and imbalance. As illustrated in Fig.~\ref{fig: Overview}(b), the SCM is under the view of data generation, where the causal and biased features are unobservable but contribute to the distribution of multimodal embeddings. 

In this paper, we propose to explicitly extract and disentangle the causal and biased features and use the learned causal features for prediction, as illustrated in Fig.~\ref{fig: Overview}(c). However, there still exists a path $\mathbf{B} \dashrightarrow \mathbf{C}  \rightarrow \mathbf{Y}$ could introduce spurious correlations. To address this issue, we further introduce a feature decorrelation mechanism to block the spurious path and enhance causal robustness.

\subsection{Dual-Stream Graph Neural Network}

Fig.~\ref{fig: Overview}(a) presents the whole architecture of the proposed dual-stream feature decorrelation framework. The core idea is to enable the model to explicitly learn causal and biased representations through parallel streams while blocking spurious correlations caused by data bias. 

We employ GRAPE~\cite{you2020handling}, a Graph Neural Network (GNN) for handling missing data in multimodal learning as the baseline. GRAPE constructs a bipartite graph $\mathcal{G} = (\mathcal{U}, \mathcal{E})$ where the node set consists of two types: patient nodes and modality nodes with one-hot encoding. Edges exist only between patient and modality nodes when modality $j$ is observed in patient $i$, where edge features are initialized as $e_{ij}^{(0)} = \text{Encoder}_j(x_j^{(i)})$ using modality-specific encoders. The framework employs multi-layer GNN for message passing. 

\subsubsection{Causal and Biased Features Extraction}
We construct two parallel GNN streams to extract causal and biased features. The two streams share the same node set $\mathcal{U}$ but dynamically adjust edge weights through an allocation mechanism. For each edge $e_{uv}^{l} \in E$ in the $l$-th layer, we design a gating function to compute assignment probabilities for each stream in the first GNN layer:
\begin{equation}
    \beta_{uv}^{(0)} = \text{MLP}_{\text{gate}}(\text{Concat}[h_u^{(0)},h_v^{(0)}]),
\end{equation}
\begin{equation}
    \tau_{uv}^{(0)} = \sigma(\beta_{uv}^{(l)}), \quad \omega_{uv}^{(0)} = 1 - \tau_{uv}^{(0)},
\end{equation}
where $\text{Concat}[\cdot]$ denotes the concatenation operator, $\sigma$ is the sigmoid function, $h_u^{(l)}$ and $h_v^{(l)}$ represent embeddings of nodes $u$ and $v$, and $\tau_{uv}^{(0)}$ and $\omega_{uv}^{(0)}$ are the weight allocations for causal and bias streams, respectively.

The edge weights are then used to modulate message propagation in each stream. The message passing in each branch follows the same structure as the baseline to generate the patient node embeddings $h_{u,c}^{(l)}$ and $h_{u,b}^{(l)}$ with adjacent nodes $v \in \mathcal{N}(u)$, but with edge-specific weight modulation:
\begin{equation}
h_{u,c}^{(l)} = U_c^{(l)} \text{Concat}(h_{u,c}^{(l-1)}, \text{Mean}(\tau_{uv}^{(l-1)} \cdot W_c^{(l)}h_{uv}^{(l-1)})),
\end{equation}
\begin{equation}
h_{u,b}^{(l)} = U_b^{(l)} \text{Concat}(h_{u,b}^{(l-1)},\text{Mean}(\omega_{uv}^{(l-1)} \cdot W_b^{(l)}h_{uv}^{(l-1)})),
\end{equation}
where $U_c^{(l)}$, $U_b^{(l)}$, $W_c^{(l)}$, and $W_b^{(l)}$ denote learnable MLP layers. 
For the first layer ($l=1$), we apply the edge-specific weights $\tau_{uv}^{(0)}$ and $\omega_{uv}^{(0)}$, 
whereas for higher layers ($l>1$), we set $\tau_{uv}^{(l-1)} = 1$ and $\omega_{uv}^{(l-1)} = 1$. 
Therefore, the two GNN streams preserve the same topology but propagate messages with complementary edge weighting schemes, allowing them to extract causal and biased information separately.

\subsubsection{Causal and Biased Features Learning}

To achieve effective disentanglement between causal and biased features, we employ a differentiated loss function design. For notational simplicity, we denote the patient-level embeddings obtained from the dual-stream GNNs as $H_c$ and $H_b$, corresponding to the causal and biased representations. Inspired by~\cite{fan2022debiasing}, the concatenated representation $E_{\text{concat}} = \text{Concat}[H_c, H_b]$ is fed into two independent classifiers: $f_c(\cdot)$ trained with standard cross-entropy loss, and $f_b(\cdot)$ with generalized cross-entropy loss to enhance bias pattern learning:
\begin{equation}
\mathcal{L}_{\text{gce}}(\psi_b(E_{\text{concat}}), y) = \frac{1 - [\psi_b(E{\text{concat}})]_y^g}{g},
\end{equation}
where parameter $g \in (0,1]$ controls the gradient amplification degree. The generalized cross-entropy loss applies larger gradients to high-confidence predictions, encouraging the bias stream to focus on extracting the biased features. The loss function is designed as:
\begin{equation}
\mathcal{L}_{\text{d}} = \mathcal{L}_{\text{ce}}(f_c(E_{\text{concat}}), y) + \mathcal{L}_{\text{gce}}(f_b(E_{\text{concat}}), y).
\end{equation}

\subsection{Feature Decorrelation Learning}

To ensure statistical independence between causal features $h_c$ and biased features $h_b$ to block the spurious correlation discussed in Sec.~\ref{sec:causal}, we propose to minimize the mutual information to decorrelate the features. The mutual information between :
\begin{equation}
I(H_c, H_b) = \int_{H_c \times H_b} \log \frac{dP_{H_c H_b}}{dP_{H_c} \otimes dP_{H_b}} dP_{H_c H_b},
\end{equation}
where $P_{H_c}$ and $P_{H_b}$ represent the marginal cumulative distribution functions of $H_c$ and $H_b$, respectively.

Since mutual information often does not have a closed-form solution, we employ the method from~\cite{belghazi2018mutual} for non-parametric estimation:
\begin{equation}
\begin{split}
\mathcal{L}_{\text{MI}} &= \sup_\omega \frac{1}{n} \sum_{i=1}^{n} \psi_\omega(H_c^i, H_b^i) \\
&\quad - \log (\frac{1}{n} \sum_{i=1}^{n} \exp \psi_\omega(H_c^i, {H_b^j}') ),
\end{split}
\end{equation}
where $\psi_\omega$ is a 2-layer MLP with parameters $\omega$, $(H_c^i, H_b^j)$ are pairs sampled from one patient, and $(H_c^i, {H_b^j}')$ are constructed pairs where ${H_b^j}'$ is obtained from samples with different labels. We sampled 10 different ${H_b^j}'$ for each pair in loss computation.

\subsection{Total Loss Function}

Combining the previous two loss functions together, our final training objective is formulated as:
\begin{equation}
L_{\text{Total}} = L_{\text{d}} + L_{\text{MI}}.
\end{equation}
The concatenated embedding $E_\textbf{concat}$ is input to classifier $f_c(\cdot)$ for inference.

\section{Experiments and results}
\subsection{Setup}
\noindent\textbf{Implementation Details.} We train the model for 100 epochs on an NVIDIA A100 GPU using the PyTorch framework. We follow the experiment protocol for fair comparison, including the batch size and learning rate. Specifically, we adopt a two-stage training strategy: first train the model with $\mathcal{L}_{\text{d}}$ for 15 epochs, then train the whole model with $\mathcal{L}_{\text{Total}}$ for the remaining 85 epochs. 

\noindent\textbf{Datasets.} We conduct comprehensive experimental evaluations on three real-world medical datasets: MIMIC-IV, eICU, and ADNI. MIMIC-IV \cite{johnson2023mimic} contains over 65,000 ICU admissions and 200,000 emergency visits from Beth Israel Deaconess Medical Center (2008-2019). eICU \cite{pollard2018eicu} includes over 200,000 ICU admissions from 208 US hospitals (2014-2015). ADNI \cite{jack2008alzheimer} is a longitudinal study of over 2,000 patients tracking Alzheimer’s disease progression. We follow the same setting as in \cite{wu2024multimodal,zhu2025causal} for experiments on the three datasets.

\begin{figure}[t]
    \centering
    \includegraphics[width=0.45\textwidth]{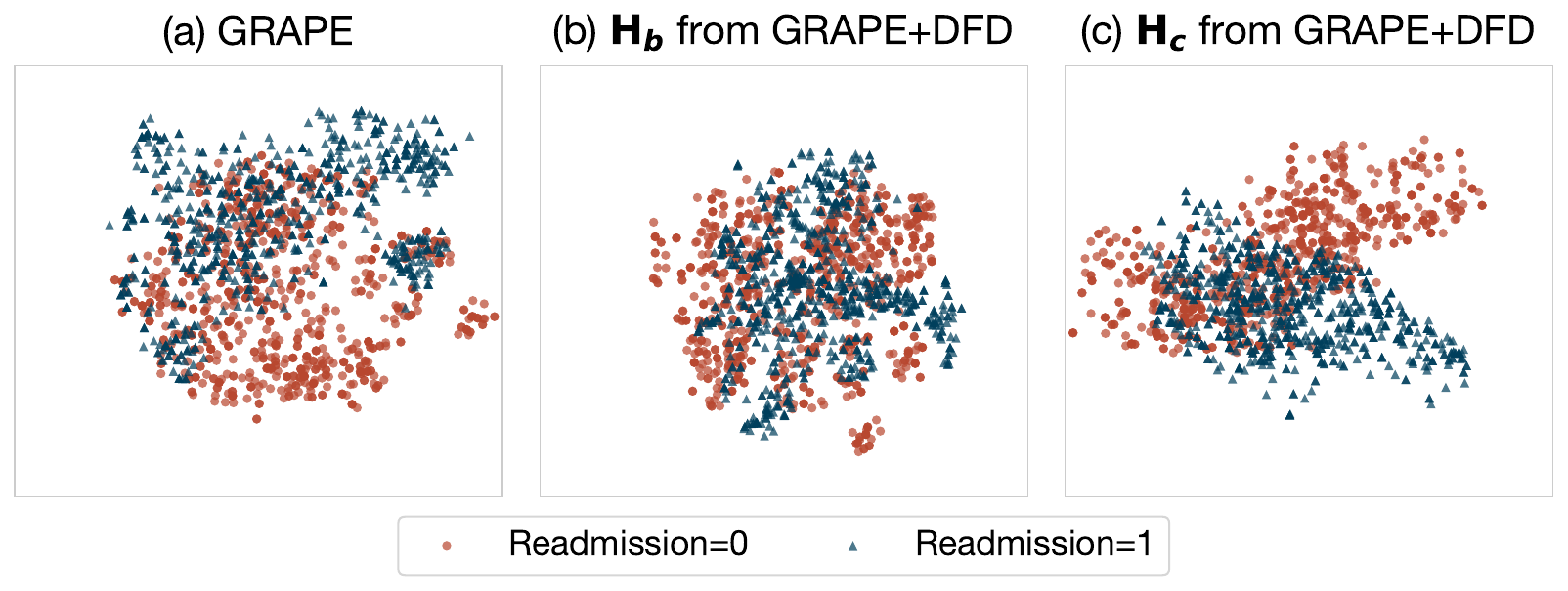}
    \caption{t-SNE visualization on MIMIC-IV readmission task.}
    \label{fig:tsne}
\end{figure}

\subsection{Main Results}
\label{mainresults}

\subsubsection{Results on the MIMIC-IV and eICU Datasets} As shown in Table~\ref{tab:icu}, we observe that MUSE+DFD achieves the new SOTA. Specifically, DFD-based MT, GRAPE, M3Care, and MUSE improve the AUC-PRC scores for mortality prediction on MIMIC-IV by 1.65\%, 2.55\%, 1.45\%, and 1.77\%, respectively. DFD also brings noticeable improvements to the eICU dataset with improvements of 1.83\%, 2.31\% and 2.23\% with GRAPE, M3Care, and MUSE in AUC-PRC for mortality prediction. These consistent improvements across different baselines demonstrate the effectiveness of integrating our framework with existing methods.

Beyond quantitative improvements, Fig.~\ref{fig:tsne} provides intuitive evidence of the effectiveness of our framework through t-SNE visualization on MIMIC-IV readmission task. Our GRAPE+DFD framework successfully disentangles representations, demonstrating the expected behavior of our dual-stream architecture. Specifically, the biased features $H_b$ remain mixed as intended, confirming they capture spurious correlations. In contrast, the causal features $H_c$ are more separated, demonstrating that our method effectively extracts discriminative causal features.

\begin{table*}[t]
    \centering
    \caption{Results on ICU datasets and ADNI. A dagger ($\dagger$) indicates the standard deviation is greater than 0.02. An asterisk (*) indicates that the DFD-based models achieves a significant improvement over the baselines, with a p-value smaller than 0.05.}
    \resizebox{0.9\textwidth}{!}{
    \renewcommand{\arraystretch}{1}
    \setlength{\tabcolsep}{2.8pt}
    \begin{tabular}{lcccccccccc}
        \toprule
        \multirow{3.7}{*}{\centering \textbf{Method}} & \multicolumn{4}{c}{\textbf{MIMIC-IV}} & \multicolumn{4}{c}{\textbf{eICU}} & \multicolumn{2}{c}{\textbf{ADNI}} \\
        \cmidrule(lr){2-5} \cmidrule(lr){6-9} \cmidrule(lr){10-11}
        & \multicolumn{2}{c}{\textbf{Mortality}} & \multicolumn{2}{c}{\textbf{Readmission}} & \multicolumn{2}{c}{\textbf{Mortality}} & \multicolumn{2}{c}{\textbf{Readmission}} & \multicolumn{2}{c}{\textbf{AD Progression}} \\
        \cmidrule(lr){2-3} \cmidrule(lr){4-5} \cmidrule(lr){6-7} \cmidrule(lr){8-9} \cmidrule(lr){10-11}
        & \textbf{AUC-ROC} & \textbf{AUC-PRC} & \textbf{AUC-ROC} & \textbf{AUC-PRC} & \textbf{AUC-ROC} & \textbf{AUC-PRC} & \textbf{AUC-ROC} & \textbf{AUC-PRC} & \textbf{AUC-ROC} & \textbf{Accuracy} \\ 
        \midrule
        CM-AE \cite{ngiam2011multimodal}   & 0.8530$^\dagger$ & 0.4351$^\dagger$ & 0.6817$^\dagger$ & 0.4324$^\dagger$ & 0.8624 & 0.3902 & 0.7462$^\dagger$ & 0.4338$^\dagger$ & 0.8722$^\dagger$ & 0.7305$^\dagger$ \\
        SMIL \cite{ma2021smil}   & 0.8607 & 0.4438 & 0.6894$^\dagger$ & 0.4368$^\dagger$ & 0.8711 & 0.4066 & 0.7506 & 0.4447 & 0.8761$^\dagger$ & 0.7338$^\dagger$ \\
        MT \cite{ma2022multimodal}    & 0.8739 & 0.4452 & 0.6901 & 0.4375 & 0.8882 & 0.4109 & 0.7635 & 0.4500 & 0.8935 & 0.7604 \\
        GRAPE \cite{you2020handling}   & 0.8837 & 0.4584$^\dagger$ & 0.7085 & 0.4551 & 0.8903 & 0.4137 & 0.7663 & 0.4501 & 0.9031$^\dagger$ & 0.7820$^\dagger$ \\
        HGMF \cite{chen2020hgmf}   & 0.8710 & 0.4433 & 0.7005$^\dagger$ & 0.4421 & 0.8878 & 0.4104 & 0.7604 & 0.4496$^\dagger$ & 0.8845$^\dagger$ & 0.7463$^\dagger$ \\
        M3Care \cite{zhang2022m3care}  & 0.8896$^\dagger$ & 0.4603$^\dagger$ & 0.7067 & 0.4532 & 0.8964 & 0.4155 & 0.7598$^\dagger$ & 0.4430 & 0.9101 & 0.7822 \\
        MUSE \cite{wu2024multimodal}  & 0.9004 & 0.4735 & 0.7152 & 0.4670$^\dagger$ & 0.9017 & 0.4216 & 0.7709 & 0.4631 & 0.9158$^\dagger$ & 0.7973$^\dagger$ \\
        \midrule
        \textbf{MT+DFD}   & 0.8833* & 0.4617* & 0.7042* & 0.4459* & 0.8877*$^\dagger$ & 0.4056* & 0.7701* & 0.4527* & 0.9057* & 0.7792* \\
        \textbf{GRAPE+DFD}   & 0.9112* & 0.4839* & 0.7373* & 0.4983* & 0.9278* & 0.4320* & 0.7799* & 0.4712* & 0.9223* & 0.8187* \\
        \textbf{M3Care+DFD}   & 0.9175* & 0.4748* & 0.7319*$^\dagger$ & 0.4967* & \textbf{0.9359*} & 0.4386* & 0.7878*$^\dagger$ & 0.4745* & 0.9292* & 0.8286* \\
        \textbf{MUSE+DFD}   & \textbf{0.9218*} & \textbf{0.4912*} & \textbf{0.7423*} & \textbf{0.5078*} & 0.9328* & \textbf{0.4439*} & \textbf{0.7966*} & \textbf{0.4858*} & \textbf{0.9275*} & \textbf{0.8278*} \\
        \bottomrule
    \end{tabular}
    }
    \label{tab:icu}
\end{table*}

\begin{table*}[t]
\centering
\caption{Ablation study on the influence of each module and variant.}
\label{tab:ablation_main}
\setlength{\tabcolsep}{2.6pt}
\renewcommand{\arraystretch}{1}
\resizebox{0.6\textwidth}{!}{
\begin{tabular}{lccccc}
\toprule
\multirow{2}{*}{\textbf{Method}} & \multicolumn{2}{c}{\textbf{MIMIC-IV}} & \multicolumn{2}{c}{\textbf{eICU}} & \multicolumn{1}{c}{\textbf{ADNI}} \\
\cmidrule(lr){2-3} \cmidrule(lr){4-5} \cmidrule(lr){6-6}
 & \textbf{Mortality} & \textbf{Readmission} & \textbf{Mortality} & \textbf{Readmission} & \textbf{AD Progression} \\
\midrule
GRAPE+DFD w/o GCE & 0.8890 & 0.7096 & 0.8973 & 0.7659 & 0.9068 \\
MUSE+DFD w/o GCE & 0.9078 & 0.7120 & 0.9046 & 0.7754 & 0.9237 \\
GRAPE+DFD w/o $\mathcal{L}_\textbf{MI}$ & 0.8950 & 0.7185 & 0.9149 & 0.7690 & 0.9098 \\
MUSE+DFD w/o $\mathcal{L}_\textbf{MI}$ & 0.9045 & 0.7208 & 0.9256 & 0.7856 & 0.9115 \\
\midrule
GRAPE+DFD & 0.9112 & 0.7373 & 0.9278 & 0.7799 & 0.9223 \\
MUSE+DFD & 0.9218 & 0.7423 & 0.9328 & 0.7966 & 0.9275\\
\bottomrule
\end{tabular}}
\label{tab:ablation}
\end{table*}

\subsubsection{Result on the ADNI Dataset} 

Table~\ref{tab:icu} reports the performance on the ADNI dataset. 
Among all methods, MUSE+DFD achieves the best results with 0.9275 AUC-ROC and 0.8278 accuracy, corresponding to relative improvements of 1.17\% and 3.05\%, respectively. GRAPE+DFD and M3Care+DFD also have substantial gains of 1.92\% and 1.91\% in AUC-ROC, further confirming the broad applicability of our approach. Importantly, these consistent gains are observed on the relatively small-scale ADNI dataset, indicating that our framework remains effective in smaller-scale settings where robust feature representation learning is essential. These results further verify the effectiveness and generalizability of our DFD framework across diverse datasets.

\subsection{Ablation Study}

We conducted comprehensive ablation studies to validate each contribution of each component. Table~\ref{tab:ablation} presents the ablation results across the datasets using AUC-ROC.

\noindent\textbf{Effectiveness of Generalized Cross-Entropy Loss.} Removing the GCE loss component significantly degrades performance across all datasets. On MIMIC-IV mortality prediction, GRAPE and MUSE show drops of 2.22\% and 1.40\% respectively, while similar degradation patterns are observed on eICU and ADNI datasets, highlighting the critical role of our causal and biased features disentanglement.

\noindent\textbf{Effect of Mutual Information Minimization.} The mutual information minimization module contributes substantial improvements across all experimental settings, with gains of 1.62\% and 1.73\% for GRAPE and MUSE on MIMIC-IV mortality prediction. The consistent performance confirm that our decorrelation mechanism successfully prevents information leakage between causal and biased streams, thereby enhancing the robustness of the learned multimodal representations.

\section{Conclusion}

This paper proposes a dual-stream feature decorrelation framework for medical multimodal representation learning from a causal perspective. Our approach employs dual-stream graph neural networks to disentangle causal and biased features, leveraging generalized cross-entropy loss and mutual information minimization to separate causal features from spurious correlations. Comprehensive experiments across MIMIC-IV, eICU, and ADNI datasets demonstrate consistent improvements. In future work, we aim to extend causal representations to medical multimodal large language models, enabling more robust reasoning in complex clinical scenarios.

\small
\bibliographystyle{IEEEbib}
\bibliography{strings,refs}

\begin{thebibliography}{10}

\bibitem{sylvain2021cmim}
Tristan Sylvain, Francis Dutil, Tess Berthier, Lisa Di~Jorio, Margaux Luck, Devon Hjelm, and Yoshua Bengio,
\newblock ``Cmim: Cross-modal information maximization for medical imaging,''
\newblock in {\em 2021-2021 IEEE International Conference on Acoustics, Speech and Signal Processing (ICASSP)}. IEEE, 2021, pp. 1190--1194.

\bibitem{johnson2023mimic}
Alistair~EW Johnson, Lucas Bulgarelli, Lu~Shen, Alvin Gayles, Ayad Shammout, Steven Horng, Tom~J Pollard, Sicheng Hao, Benjamin Moody, Brian Gow, et~al.,
\newblock ``Mimic-iv, a freely accessible electronic health record dataset,''
\newblock {\em Sci. Data}, vol. 10, no. 1, pp. 1, 2023.

\bibitem{wagner2020ptb}
Patrick Wagner, Nils Strodthoff, Ralf-Dieter Bousseljot, Dieter Kreiseler, Fatima~I Lunze, Wojciech Samek, and Tobias Schaeffter,
\newblock ``Ptb-xl, a large publicly available electrocardiography dataset,''
\newblock {\em Sci. Data}, vol. 7, no. 1, pp. 1--15, 2020.

\bibitem{maltesen2020longitudinal}
Raluca~Georgiana Maltesen, Reinhard Wimmer, and Bodil~Steen Rasmussen,
\newblock ``A longitudinal serum nmr-based metabolomics dataset of ischemia-reperfusion injury in adult cardiac surgery,''
\newblock {\em Sci. Data}, vol. 7, no. 1, pp. 198, 2020.

\bibitem{allen2014uk}
Naomi~E Allen, Cathie Sudlow, Tim Peakman, Rory Collins, and Uk~biobank,
\newblock ``Uk biobank data: come and get it,'' 2014.

\bibitem{liu2025multimodal}
Jing Liu, Linxiao Gong, Juncen Guo, Jingyi Wu, Lianlong Sun, Yulai Bi, Kartik Patwari, Boan Chen, Lichi Zhang, Wei Zhou, et~al.,
\newblock ``Multimodal large language models in medicine and nursing: A survey,''
\newblock {\em Authorea Preprints}, 2025.

\bibitem{su2025large}
Xiu Su, Qinghua Mao, Zhongze Wu, Xi~Lin, Shan You, Yue Liao, and Chang Xu,
\newblock ``Large language models driven neural architecture search for universal and lightweight disease diagnosis on histopathology slide images,''
\newblock {\em npj Digital Medicine}, vol. 8, no. 1, pp. 682, 2025.

\bibitem{ma2022multimodal}
Mengmeng Ma, Jian Ren, Long Zhao, Davide Testuggine, and Xi~Peng,
\newblock ``Are multimodal transformers robust to missing modality?,''
\newblock in {\em CVPR}, 2022, pp. 18177--18186.

\bibitem{chen2020hgmf}
Jiayi Chen and Aidong Zhang,
\newblock ``Hgmf: heterogeneous graph-based fusion for multimodal data with incompleteness,''
\newblock in {\em KDD}, 2020, pp. 1295--1305.

\bibitem{zhang2022m3care}
Chaohe Zhang, Xu~Chu, Liantao Ma, Yinghao Zhu, Yasha Wang, Jiangtao Wang, and Junfeng Zhao,
\newblock ``M3care: Learning with missing modalities in multimodal healthcare data,''
\newblock in {\em KDD}, 2022, pp. 2418--2428.

\bibitem{wu2024multimodal}
Zhenbang Wu, Anant Dadu, Nicholas Tustison, Brian Avants, Mike Nalls, Jimeng Sun, and Faraz Faghri,
\newblock ``Multimodal patient representation learning with missing modalities and labels,''
\newblock in {\em ICLR}, 2024.

\bibitem{LIU2022109697}
Chuan Liu, Jingwei Wang, Yunkang Cao, Min Liu, and Weiming Shen,
\newblock ``Gon: End-to-end optimization framework for constraint graph optimization problems,''
\newblock {\em Knowledge-Based Systems}, vol. 254, pp. 109697, 2022.

\bibitem{10219171}
Jingwei Wang, Chuan Liu, Yukai Zhao, Zhirui Zhao, Yunlong Ma, Min Liu, and Weiming Shen,
\newblock ``Graph convolutional network aided inverse graph partitioning for resource allocation,''
\newblock {\em IEEE Trans. Indus. Infor.}, vol. 20, no. 3, pp. 3082--3091, 2024.

\bibitem{cross2024bias}
James~L Cross, Michael~A Choma, and John~A Onofrey,
\newblock ``Bias in medical ai: Implications for clinical decision-making,''
\newblock {\em PLOS Digital Health}, vol. 3, no. 11, pp. e0000651, 2024.

\bibitem{norori2021addressing}
Natalia Norori, Qiyang Hu, Florence~Marcelle Aellen, Francesca~Dalia Faraci, and Athina Tzovara,
\newblock ``Addressing bias in big data and ai for health care: A call for open science,''
\newblock {\em Patterns}, vol. 2, no. 10, 2021.

\bibitem{deshpande2022deep}
Shachi Deshpande, Kaiwen Wang, Dhruv Sreenivas, Zheng Li, and Volodymyr Kuleshov,
\newblock ``Deep multi-modal structural equations for causal effect estimation with unstructured proxies,''
\newblock {\em NeurIPS}, vol. 35, pp. 10931--10944, 2022.

\bibitem{yang2021causalvae}
Mengyue Yang, Furui Liu, Zhitang Chen, Xinwei Shen, Jianye Hao, and Jun Wang,
\newblock ``Causalvae: Disentangled representation learning via neural structural causal models,''
\newblock in {\em CVPR}, 2021, pp. 9593--9602.

\bibitem{liu2025privacy}
Yang Liu, Siao Liu, Xiaoguang Zhu, Jielin Li, Hao Yang, Liangyu Teng, Juncen Guo, Yan Wang, Dingkang Yang, and Jing Liu,
\newblock ``Privacy-preserving video anomaly detection: A survey,''
\newblock {\em IEEE Transactions on Neural Networks and Learning Systems}, pp. 1--22, 2025.

\bibitem{pearl2016causal}
Judea Pearl, Madelyn Glymour, and Nicholas~P Jewell,
\newblock {\em Causal inference in statistics: A primer},
\newblock John Wiley \& Sons, 2016.

\bibitem{pearl2009causality}
Judea Pearl,
\newblock {\em Causality},
\newblock Cambridge university press, 2009.

\bibitem{liu2025crcl}
Yang Liu, Hongjin Wang, Zepu Wang, Xiaoguang Zhu, Jing Liu, Peng Sun, Rui Tang, Jianwei Du, Victor~CM Leung, and Liang Song,
\newblock ``Crcl: Causal representation consistency learning for anomaly detection in surveillance videos,''
\newblock {\em IEEE Trans. Image Process.}, 2025.

\bibitem{you2020handling}
Jiaxuan You, Xiaobai Ma, Yi~Ding, Mykel~J Kochenderfer, and Jure Leskovec,
\newblock ``Handling missing data with graph representation learning,''
\newblock {\em NeurIPS}, vol. 33, pp. 19075--19087, 2020.

\bibitem{fan2022debiasing}
Shaohua Fan, Xiao Wang, Yanhu Mo, Chuan Shi, and Jian Tang,
\newblock ``Debiasing graph neural networks via learning disentangled causal substructure,''
\newblock {\em NeurIPS}, vol. 35, pp. 24934--24946, 2022.

\bibitem{belghazi2018mutual}
Mohamed~Ishmael Belghazi, Aristide Baratin, Sai Rajeshwar, Sherjil Ozair, Yoshua Bengio, Aaron Courville, and Devon Hjelm,
\newblock ``Mutual information neural estimation,''
\newblock in {\em International conference on machine learning}. PMLR, 2018, pp. 531--540.

\bibitem{pollard2018eicu}
Tom~J Pollard, Alistair~EW Johnson, Jesse~D Raffa, Leo~A Celi, Roger~G Mark, and Omar Badawi,
\newblock ``The eicu collaborative research database, a freely available multi-center database for critical care research,''
\newblock {\em Sci. Data}, vol. 5, no. 1, pp. 1--13, 2018.

\bibitem{jack2008alzheimer}
Clifford~R Jack~Jr, Matt~A Bernstein, Nick~C Fox, Paul Thompson, Gene Alexander, Danielle Harvey, Bret Borowski, Paula~J Britson, Jennifer L.~Whitwell, Chadwick Ward, et~al.,
\newblock ``The alzheimer's disease neuroimaging initiative (adni): Mri methods,''
\newblock {\em J. Magn. Reson. Imaging}, vol. 27, no. 4, pp. 685--691, 2008.

\bibitem{zhu2025causal}
Xiaoguang Zhu, Lianlong Sun, Yang Liu, Pengyi Jiang, Uma Srivatsa, Nipavan Chiamvimonvat, and Vladimir Filkov,
\newblock ``Causal debiasing medical multimodal representation learning with missing modalities,''
\newblock {\em arXiv preprint arXiv:2509.05615}, 2025.

\bibitem{ngiam2011multimodal}
Jiquan Ngiam, Aditya Khosla, Mingyu Kim, Juhan Nam, Honglak Lee, Andrew~Y Ng, et~al.,
\newblock ``Multimodal deep learning.,''
\newblock in {\em ICML}, 2011, vol.~11, pp. 689--696.

\bibitem{ma2021smil}
Mengmeng Ma, Jian Ren, Long Zhao, Sergey Tulyakov, Cathy Wu, and Xi~Peng,
\newblock ``Smil: Multimodal learning with severely missing modality,''
\newblock in {\em AAAI}, 2021, vol.~35, pp. 2302--2310.

\end{thebibliography}

\end{document}